%% file: main.tex
\documentclass[10pt,twocolumn,letterpaper]{article}

\usepackage{cvpr}              

\usepackage{graphicx}
\usepackage{amsmath}
\usepackage{amssymb}
\usepackage{booktabs}
\usepackage{multirow}
\usepackage[accsupp]{axessibility}

\usepackage[pagebackref,breaklinks,colorlinks]{hyperref}

\usepackage[capitalize]{cleveref}
\crefname{section}{Sec.}{Secs.}
\Crefname{section}{Section}{Sections}
\Crefname{table}{Table}{Tables}
\crefname{table}{Tab.}{Tabs.}

\def\cvprPaperID{5983} 
\def\confName{CVPR}
\def\confYear{2022}

\begin{document}

\title{Beyond Supervised vs. Unsupervised: Representative Benchmarking and Analysis of Image Representation Learning}

\author{Matthew Gwilliam and Abhinav Shrivastava\\
University of Maryland, College Park\\
}
\maketitle

\begin{abstract}
   By leveraging contrastive learning, clustering, and other pretext tasks, unsupervised methods for learning image representations have reached impressive results on standard benchmarks.
   The result has been a crowded field -- many methods with substantially different implementations yield results that seem nearly identical on popular benchmarks, such as linear evaluation on ImageNet.
   However, a single result does not tell the whole story.
   In this paper, we compare methods using performance-based benchmarks such as linear evaluation, nearest neighbor classification, and clustering for several different datasets, demonstrating the lack of a clear front-runner within the current state-of-the-art.
   In contrast to prior work that performs only supervised vs. unsupervised comparison, we compare several different unsupervised methods against each other.
   To enrich this comparison, we analyze embeddings with measurements such as uniformity, tolerance, and centered kernel alignment (CKA), and propose two new metrics of our own: nearest neighbor graph similarity and linear prediction overlap.
   We reveal through our analysis that in isolation, single popular methods should not be treated as though they represent the field as a whole, and that future work ought to consider how to leverage the complimentary nature of these methods.
   We also leverage CKA to provide a framework to robustly quantify augmentation invariance, and provide a reminder that certain types of invariance will be undesirable for downstream tasks.
\end{abstract}

\input{sections/1-introduction}

\input{sections/2-background}

\input{sections/3-methods}

\input{sections/4-analysis}

\input{sections/5-conclusion}

{\small
\bibliographystyle{ieee_fullname}
\bibliography{main}
}

\newpage

\input{sections/appendix}

\end{document}

%% file: sections/1-introduction.tex
\section{Introduction}
\label{sec:introduction}

\begin{figure}
    \centering
    \includegraphics[width=\linewidth]{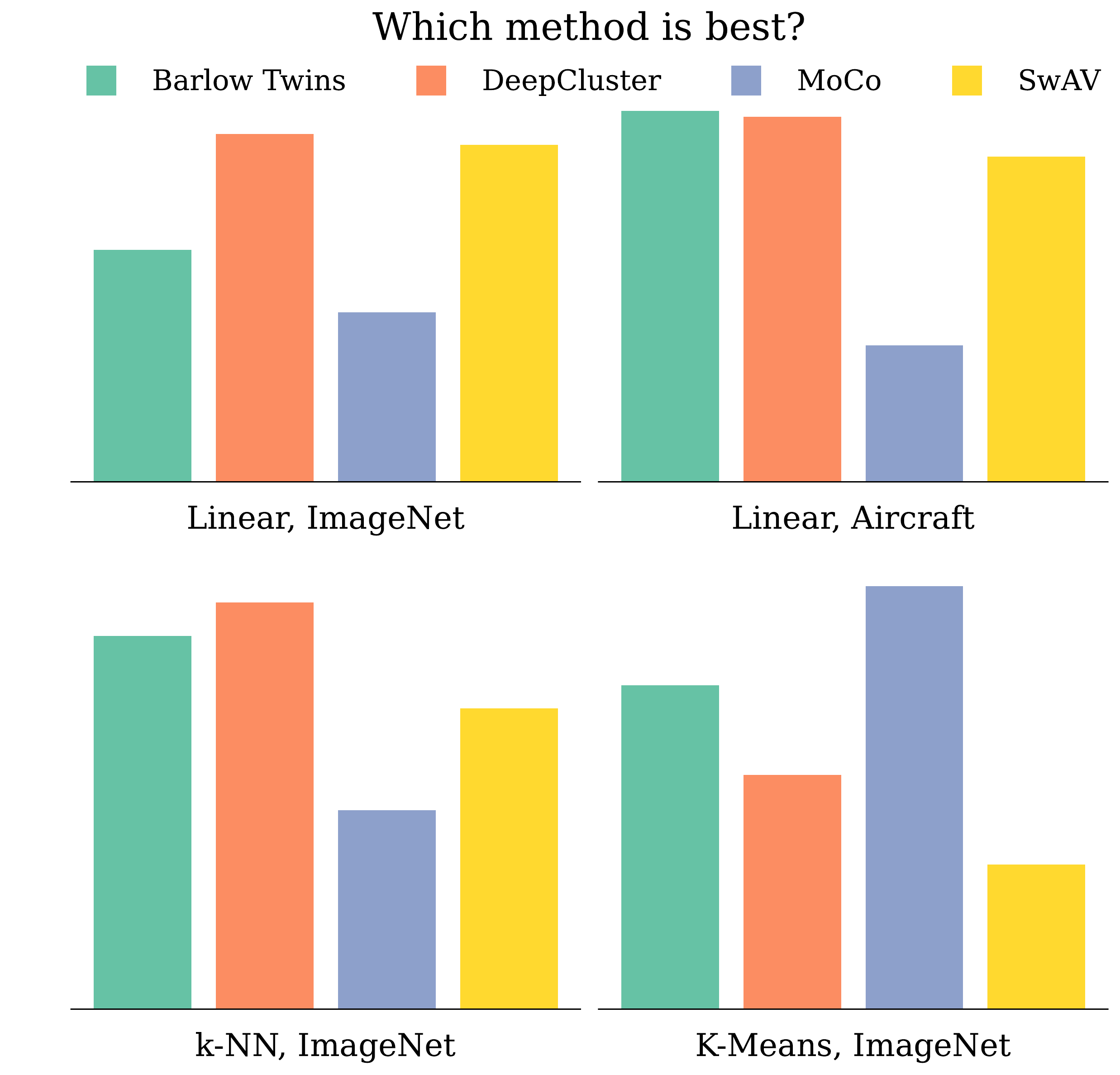}
    \caption{Results for a sample of classification benchmarks we perform in this paper. While these bar charts report real results, lack of axes is intentional -- the exact numbers are in Section~\ref{sec:analysis}. Importantly, between the four tasks, there is no clear ``best'' method.}
    \label{fig:teaser_square}
\end{figure}

\begin{figure*}
    \centering
    \includegraphics[width=\linewidth]{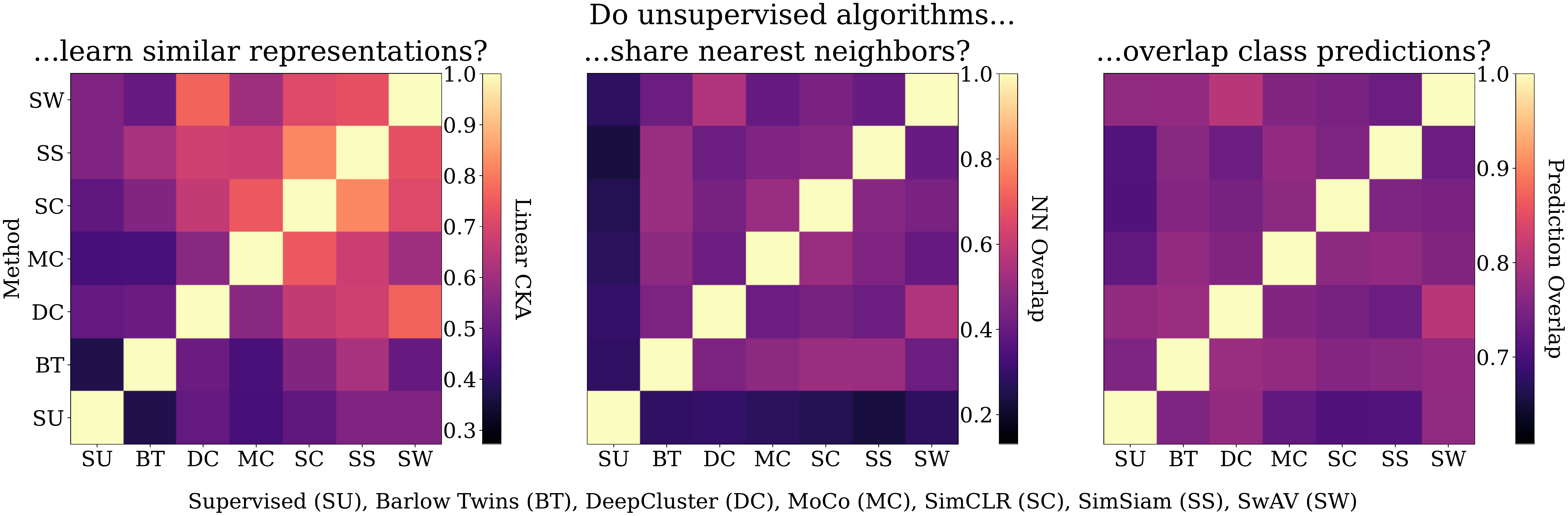}
    \caption{Similarity between learned representations, based on the outputs of ResNet50s on the validation images from ImageNet. For the metrics shown, which are described in more detail in Section~\ref{sec:analysis}, higher values indicate similarity. While the supervised model tends to be more dissimilar from the unsupervised models, there are many ways in which unsupervised methods differ substantially from each other.}
    \label{fig:teaser_star}
    
\end{figure*}

Image features are critical components in many computer vision (CV) pipelines. 
In this paper, we define image features, also referred to as embeddings, encodings, or representations, as an $n$-dimensional vector that represents the content of an image.
With the emergence of deep learning, classical approaches to computing image features have been supplanted by neural networks that use large amounts of data to generate powerful image representations.
The most widespread method is straightforward: a neural network (\eg, a ResNet50~\cite{he2015deep}) is trained to classify the images in some large dataset, typically ImageNet.
The portion of the network that performs the classification, usually just the final layer, is then removed, and the outputs of the penultimate layer for a given image are considered the features for that image.
This process relies on image classification, a supervised learning task, and thus requires the availability of large amounts of annotated, high-quality data.

Recent successes make unsupervised learning a viable alternative paradigm where image features are learned without the need for class labels.
Within unsupervised learning, methods can be considered either generative or discriminative.
Generative methods are typically designed for reconstruction or similar tasks~\cite{brock2018large,donahue2016adversarial,donahue2019large,kingma2013auto,vincent2008extracting}.
Since we are more concerned with a potential transfer to downstream tasks such as image classification and object detection, we choose to focus on discriminative methods.

There are many different ways to compare image representation learning algorithms.
In this paper, we opt to focus on the role of the methods as feature extractors, where a model that is pre-trained for some task is expected to be able to generate useful features for unseen images.
Thus, we only use benchmarks that keep the backbone (the portion of the neural network that generates the embedding) frozen.
Prior works are often limited by their focus on a single benchmark, single method, or toy datasets.
In contrast, we compare 6 SOTA unsupervised methods on ImageNet and 6 fine-grained visual categorization (FGVC) datasets using several different benchmarks.
Figure~\ref{fig:teaser_square} provides a sample of this angle of analysis.

Comparing these methods to each other is very important.
Nevertheless, prior analysis works tend to lump unsupervised methods together, and often choose only a single representative such as MoCo or SimCLR for comparison against supervised representation learning~\cite{DBLP:journals/corr/abs-2012-00868,DBLP:journals/corr/abs-2005-10242,DBLP:journals/corr/abs-2106-05967,DBLP:journals/corr/abs-2105-05837,grigg2021selfsupervised}.
This ignores the significant ways in which unsupervised methods differ from each other.
In contrast to prior work, we extend existing methods and introduce novel methods to prove that unsupervised methods vary significantly in terms of how they learn to represent images, as shown in Figure~\ref{fig:teaser_star}.

State-of-the-art convolution-based unsupervised algorithms, whether they use contrastive learning, clustering, or some pretext task such as colorization, all attempt to learn invariance to some class of augmentations.
In other words, they seek to learn a function $f$, such that $f(I) = f(I_A)$ for some image $I$ and some set of augmentations $A$ that are applied to that image.
Xiao \etal speculate on the negative effect this may have on learned representations and performance on downstream tasks~\cite{xiao2021what}.
However, a careful reading reveals they don't provide evidence for the existence of transform invariance in unsupervised models, only that their method seems to perform better than MoCo on tasks related to transform invariance.
This inspires Section~\ref{subsec:augmentation-invariance-results} -- 
we take a closer look at the presence of augmentation invariance in representations learned by different unsupervised methods.

Prior work is constrained by some combination of limited metrics, use of toy datasets, and a tendency to consider a single unsupervised method as though it is representative of the field. In contrast to this, we contribute the following:
\begin{itemize}
    \item We utilize multiple methods for measuring properties of learned embeddings, including 3 performance-based benchmarks, and an extension of prior work on uniformity-tolerance analysis to more unsupervised methods across more realistic datasets.
    \vspace{-0.5em}
    \item We perform novel comparison by extending Linear Centered Kernel Alignment (CKA) analysis beyond toy datasets, and by developing two new metrics for comparing embeddings: nearest neighbor graph similarity and linear overlap.
    \vspace{-0.5em}
    \item We propose a framework for measuring augmentation invariance, and demonstrate its results across several methods, augmentations, and datasets.
    \vspace{-0.5em}
\end{itemize}

\noindent{We conclude in Section~\ref{sec:conclusion} with key insights for future unsupervised methods for representation learning:}
\begin{itemize}
    \item Currently, there is no clear ``best'' method.
    \vspace{-0.5em}
    \item Unsupervised models share properties that are circumstantially undesirable, \eg, color invariance.
    \vspace{-0.5em}
    \item Unsupervised models have similar representations in most layers, but diverge substantially in the last layer.
    \vspace{-0.5em}
\end{itemize}

%% file: sections/2-background.tex
\section{Related Work}
\label{sec:background}

\subsection{Unsupervised Learning Methods}

Some of the first unsupervised methods of the deep learning era were fashioned after pretext tasks from natural language processing. A network would be trained to perform some auxiliary task before transfer for downstream tasks. 
These auxiliary pretexts task included solving jigsaw puzzles~\cite{noroozi2016unsupervised}, colorization from grayscale~\cite{larsson2016learning,zhang2016colorful}, inpainting~\cite{pathak2016context}, relative patch prediction~\cite{doersch2015unsupervised}, predicting rotation angle~\cite{gidaris2018unsupervised}, or a combination of tasks~\cite{doersch2017multi}.
However, the introduction of Noise Contrastive Estimation (NCE)~\cite{gutmann2010noise} triggered a paradigm shift within unsupervised learning~\cite{Wu_2018_CVPR}, and subsequent methods which utilized contrastive learning~\cite{pmlr-v119-chen20j,chen2020improved,misra2020self} would surpass all of these.

\textbf{Contrastive Learning}, which implicitly performs instance discrimination, involves training a model to attract positive pairs (typically augmented views of a given image) and repel negative pairs (augmented views of two different images)~\cite{gutmann2010noise,hadsell2006dimensionality}. Many papers have proposed successful methods using contrastive learning~\cite{bachman2019learning,chen2020simple,chen2020improved,DBLP:journals/corr/abs-1911-05722,hjelm2018learning,henaff2020data,misra2020self,oord2018representation,tian2020contrastive,Wu_2018_CVPR,Ye_2019_CVPR}. In this study, we include Barlow Twins~\cite{DBLP:journals/corr/abs-2103-03230}, MoCov2~\cite{chen2020improved}, and SimCLR~\cite{chen2020simple}.

\textbf{Clustering} has emerged as another important class of unsupervised methods~\cite{asano2019self,caron2018deep,caron2019unsupervised,caron2020unsupervised}. Popular methods such as DeepCluster~\cite{caron2018deep} and SwAV~\cite{caron2020unsupervised}, and even methods that don't explicitly attempt representation learning, such as SCAN~\cite{van2020scan}, share many traits with contrastive learning. Among these are the tendency to rely on a large batch size, the use of strong augmentations, which all of these have in common, and implementation details such as the use of projection heads, which are used by SwAV and DeepClusterv2~\cite{caron2018deep,caron2020unsupervised}. We take as our representatives from this category DeepClusterv2 and SwAV. 

\textbf{Other} methods, such as SimSiam~\cite{chen2021exploring}, use neither negative pairs nor clustering objectives. 
Other methods highlight the potential of vision transformers within unsupervised regimes~\cite{caron2021emerging,li2021efficient,xie2021self}. 
However, in this paper, we choose to compare methods that use ResNet-50 backbones. 
Furthermore, we believe the sample of methods we select are sufficient to support our main points.

\subsection{Analysis of Unsupervised Learning} 

While each paper proposing a new algorithm uses some tasks to attempt to demonstrate their success compared to the prior art, other popular papers have entirely focused on benchmarking, evaluation, and comparison of specific methods. \cite{DBLP:journals/corr/abs-2007-13916} studies augmentation invariance from the perspective of accuracy by using natural images that attempt to vary certain conditions, such as illumination. \cite{DBLP:journals/corr/abs-2005-10242} and \cite{Wang_2021_CVPR} address properties of the learned embeddings such as alignment, uniformity, and tolerance. Other works benchmark unsupervised performance on various tasks and conditions~\cite{DBLP:journals/corr/abs-2106-05967,DBLP:journals/corr/abs-2105-05837,DBLP:journals/corr/abs-2103-16483,Kotar_2021_ICCV}. \cite{grigg2021selfsupervised} uses the centered kernel alignment (CKA) framework from \cite{pmlr-v97-kornblith19a} to compare supervised and unsupervised representations. 

This prior work operates under certain constraints.
Many papers consider only a single unsupervised method, either MoCo~\cite{DBLP:journals/corr/abs-2012-00868,DBLP:journals/corr/abs-2005-10242,DBLP:journals/corr/abs-2106-05967} or SimCLR~\cite{DBLP:journals/corr/abs-2105-05837,grigg2021selfsupervised}, as though it represents the entire field; of those that consider other methods, none consider more than three~\cite{DBLP:journals/corr/abs-2007-13916,DBLP:journals/corr/abs-2103-16483,Kotar_2021_ICCV}.
We extend uniformity-tolerance analysis~\cite{Wang_2021_CVPR} to multiple unsupervised methods. 
We extend CKA analysis~\cite{JMLR:v13:cortes12a,pmlr-v97-kornblith19a,grigg2021selfsupervised} beyond tiny datasets, and to multiple methods. 
We perform FGVC benchmarking~\cite{DBLP:journals/corr/abs-2105-05837,DBLP:journals/corr/abs-2103-16483} for additional metrics and algorithms.
We perform CKA analysis~\cite{pmlr-v97-kornblith19a} to examine augmentation invariance without relying on linear classification as a confounding intermediate step~\cite{DBLP:journals/corr/abs-2007-13916}.
We develop additional methods for comparing pairs or groups of methods for our first-of-its-kind comprehensive analysis of the similarities and complementary attributes of pretrained unsupervised methods for image representation learning.
Our work follows in the spirit of Ericcson et al. in that we perform a comprehensive analysis using a sample of many unsupervised methods~\cite{DBLP:journals/corr/abs-2011-13377}.
However, by using an entirely different set of tasks and datasets, we are able to both provide further evidence for one of their main conclusions (that no one method is \textit{the best}), and uncover novel insights as well.











%% file: sections/3-methods.tex
\section{Methods}
\label{sec:methods}

\subsection{Performance-based Comparison}\label{subsec:performance-comparison}

To measure the quality of the learned representations, we perform three performance-based measurements. 
For linear evaluation, we use the VISSL repository~\cite{goyal2021vissl} to train a linear classifier on frozen features. 
For $k$-nearest neighbor classification, we also use the settings in VISSL, varying the number of neighbors for fine-grained datasets as described in the appendix. 
For $k$-means clustering, we try 10 initializations with the k-means++ method~\cite{arthur2006k}, and for ImageNet we use mini-batches of 16,384 images.

To obtain the accuracy for k-means cluster assignments when the number of classes is equal to the number of clusters, we use hungarian matching, mapping clusters to classes one-to-one in such a way that maximizes correspondence to ground truth classes.
For overclustering, we greedily map each cluster to the ground truth class which has the most images in the cluster.
Accuracy is then computed normally, using the cluster-to-class mapping result as the prediction.
For each of linear evaluation, k-NN classification, and k-means clustering, models are trained on training data, and results come from the evaluation on test data.  

\subsection{Uniformity-Tolerance Tradeoff}

To analyze how embeddings are distributed on the hypersphere, we borrow two key properties from prior work: uniformity~\cite{DBLP:journals/corr/abs-2005-10242} and tolerance~\cite{Wang_2021_CVPR}.
Uniformity, $U$, which describes how closely the embeddings match a uniform distribution on a hypersphere, is defined in Equation~\ref{eq:uniformity}, where $t$ is a scaling hyperparameter that we set to $2$, $f$ represents the model, and $x$ and $y$ are any pair of images. 

\begin{equation}\label{eq:uniformity}
    U = \text{log} \mathop{\mathbb{E}}_{x,y \sim p_\text{data}}\left[e^{-t||f(x)-f(y)||_2^2}\right]
\end{equation}

Tolerance, $T$, is given in Equation~\ref{eq:tolerance}, where $f$, $x$, and $y$ are the same as in Equation~\ref{eq:uniformity}, and $I$ is the indicator function for the ground truth labels, returning $1$ when $x$ and $y$ belong to the same class, and $0$ otherwise.


\begin{equation}\label{eq:tolerance}
    T =  \mathop{\mathbb{E}}_{x,y \sim p_\text{data}}\left[ (||f(x)||_2^T ||f(y)||_2) \cdot I_{l(x)=l(y)} \right]
\end{equation}

Whereas uniformity measures how equally spread out the features are, tolerance leverages ground truth labels to indicate how well the embeddings reflect the semantic relationships between the images. 

\begin{figure*}
    \centering
    \includegraphics[width=\linewidth]{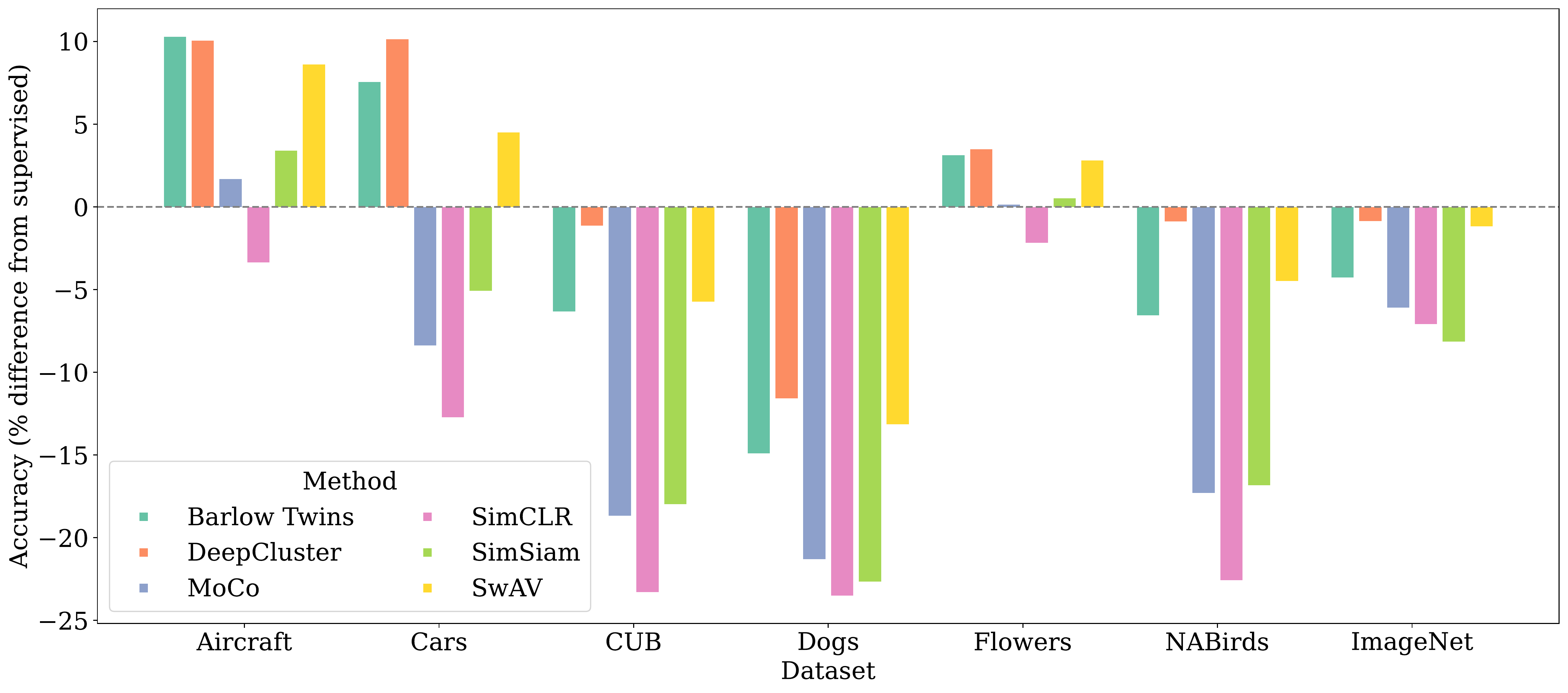}
    \caption{\textbf{Linear Classifier Results} on ImageNet and 6 FGVC datasets. 
    Barlow Twins, DeepCluster, and SwAV tend to outperform Moco, SimCLR, and SimSiam, but there is no obvious winner.}
    \label{fig:all_frozen_results}
    \vspace{-0.1in}
\end{figure*}

\subsection{Linear CKA for Comparing Representations}

We follow procedures from prior work to compute Centered Kernel Alignment (CKA) values~\cite{pmlr-v97-kornblith19a,grigg2021selfsupervised}, including using only a linear kernel~\cite{grigg2021selfsupervised}. 
To compute this, we first obtain the matrices containing the embeddings for two different methods, such as SimCLR and MoCo, which we represent $X$ and $Y$. 
We then compute the Gram matrices of the embedding matrices: $K$ = $X X^T$, $L$ = $Y Y^T$. The CKA value is given by the normalized Hilbert-Schmidt Independence Criterion (HSIC)~\cite{gretton2007kernel} as follows:

\begin{equation}
    \text{CKA}(K,L) = \frac{\text{HSIC}(K,L)}{\sqrt{\text{HSIC}(K,K) \text{HSIC}(L,L)}}
\end{equation}

Prior work~\cite{grigg2021selfsupervised} performs these computations for tiny datasets, consisting of $32 \times 32$ images. 
Since we extend this to $224 \times 224$ images, we compensate for the increased memory requirements by taking a set random sample of 10,000 test images when working with ImageNet.

\subsection{Proposed Metric: NN Graph Similarity}\label{subsec:nn-similarity}

Even for pure contrastive learning, images that belong to the same ground truth class tend to be semantically similar.
In contrast to this tolerance, which relies on ground truth labels and describes the semantic structure in terms of a single model, we propose an unsupervised way to compare the structure of the semantic relationships between two or more learned representations.
Specifically, we consider models in terms of their nearest neighbor graphs for a given dataset.
Each image is a node, and an image's top-$k$ neighbors are represented by directed edges.
We can thus compute the similarity between two representations by comparing their nearest neighbor graphs.
We choose to perform this computation in terms of neighbor overlap, where neighbor overlap refers to the average number of shared edges (neighbors) per node (image) for the graphs (unsupervised algorithms) considered. 
This neighbor overlap conveys the similarity in semantic structure learned by different algorithms. A score of 1.0 would indicate the structures are identical -- the images have the same nearest $k$ neighbors for both unsupervised algorithms.
A score of 0.0 would indicate there are no shared neighbors.
We use this method to compare pairs of models, as in Figure~\ref{fig:teaser_star}.

\subsection{Proposed Metric: Linear Prediction Overlap}

Unlike nearest neighbor graph similarity, this metric takes an indirect approach to compare two or more representations.
For each image in a dataset, we get the predictions from the linear classifiers trained on each unsupervised backbone, described in Section~\ref{subsec:performance-comparison}.
We then perform a few different calculations.
For analysis that relies on ground truth labels, we calculate the portion of the dataset that all models classify correctly, or that no models classify correctly, or that only some subset of models from a group classifies correctly.
For analysis that ignores the labels, we compute the percentage of the dataset for which some set of classifiers has the same prediction, regardless of the correctness of the prediction.
For this measure, a score of 1.0 would indicate the set of classifiers make the same predictions for all images, while 0.0 would mean they do not have identical predictions for any image.
We thus use linear overlap both to compare pairs of methods, as in Figure~\ref{fig:teaser_star}, as well as to compare sets of multiple models, as in Table~\ref{tab:linear-overlap-results}.

\subsection{Proposed Analysis: Augmentation Invariance}\label{subsec:analyzing-augmentation-invariance}

Popular unsupervised algorithms train models to be augmentation invariant. 
We develop a method for analyzing the prevalence of augmentation invariance in the representations learned by unsupervised models.
Unlike prior work~\cite{DBLP:journals/corr/abs-2007-13916}, we consider a broader set of unsupervised algorithms, and perform measurements on learned representations directly rather than relying on the performance of learned linear classifiers as a proxy.
Instead, we use CKA to compare the similarity between embeddings of augmented and non-augmented images.
We take $\text{CKA}(K,L)$ on $K$ = $X X^T$, $L$ = $X_A X_A^T$ where $X$ is the embedding matrix for the images for a given dataset, and $X_A$ is the embedding matrix for the same images after some augmentation $A$. 
We thus extend Linear CKA for unsupervised algorithms to overcome the limitations of previous methods and look directly at augmentation invariance for image representations.

%% file: sections/4-analysis.tex
\section{Analysis}
\label{sec:analysis}

For our representative set of discriminative unsupervised methods, we consider contrastive methods SimCLR~\cite{chen2020simple} and MoCo~\cite{chen2020improved}, clustering methods DeepClusterv2~\cite{caron2018deep} and SwAV~\cite{caron2020unsupervised}, as well as Barlow Twins~\cite{DBLP:journals/corr/abs-2103-03230}, which attempts redundancy reduction, and SimSiam~\cite{chen2021exploring}, which uses neither negative pairs nor clustering.
For SimCLR, DeepCluster, and SwAV, we use the 800 epoch checkpoints from the VISSL model zoo~\cite{goyal2021vissl}, unless otherwise specified.
For Barlow Twins, we use VISSL's 1000 epoch checkpoint.
For MoCo, we use the 800 epoch checkpoint from the authors.
For SimSiam, we use the author-provided 100 epoch checkpoint.
While we could have opted to retrain the models, we believe that discrepancy in training time is not a legitimate confounding factor in our analysis, and any attempt to create some ``fair'' setting would inevitably favor whichever models perform best under that setting.
These methods can be tuned on a variety of hyperparameters, and any given setting that is ``fair'' to one inevitably favors others, so we opt to use the settings for which models are available.
Additionally, our ablation experiments (Figure~\ref{fig:cka_ablation}) indicate training time would not impact any of our findings.

We perform the experiments here on some subset of the datasets in Table~\ref{tab:datasets}, which contain realistic images.

\subsection{Performance Benchmarks}

As explained in Section~\ref{sec:methods}, we perform VISSL's linear evaluation, where we train linear classifiers on frozen features from the first convolutional layer and for each of the 4 bottleneck blocks of the ResNet-50 network.
We show results for the linear classifier trained on the outputs of the final block in Figure~\ref{fig:all_frozen_results}. 
We show results for k-NN classification in Table~\ref{tab:knn-results} and for k-means clustering in Table~\ref{tab:k-means-results}.
As an additional benchmark for DeepCluster and SwAV, we also compare their clustering heads, which partition ImageNet into 3000 clusters, to k-means clustering in Table~\ref{tab:k-means-overcluster-results}.
We don't consider other performance benchmarks such as full finetuning because we are trying to evaluate only the learned embeddings, and not the network initializations.

\begin{table}[ht]
    \centering
    {
     \setlength{\tabcolsep}{0.3em}
    \caption{\textbf{Datasets} used for experiments in this paper.}
    \label{tab:datasets}
    \vspace{-0.1in}
    \begin{tabular}{@{}l l c c c@{}} 
        \toprule
        Dataset & {} & \multicolumn{1}{c}{\#Cls} & \multicolumn{1}{c}{\#Train} & \multicolumn{1}{c}{\#Test} \\
        \midrule
        FGVC Aircraft \cite{maji13fine-grained} & (Aircraft) & 100 & 6,667 & 3,333 \\
        Stanford Cars \cite{KrauseStarkDengFei-Fei_3DRR2013} & (Cars) & 196 & 8,144 & 8,041 \\
        Caltech Birds \cite{WahCUB_200_2011} & (CUB) & 200 & 5,994 & 5,794 \\
        Stanford Dogs \cite{KhoslaYaoJayadevaprakashFeiFei_FGVC2011} & (Dogs) & 120 & 12,000 & 8,580 \\
        Oxford Flowers \cite{Nilsback08} & (Flowers) & 102 & 2,040 & 6,149 \\
        NABirds V1 \cite{7298658} & (NABirds) & 555 & 23,929 & 24,633 \\
        ImageNet \cite{5206848} & & 1000 & 1.3mil & 50,000 \\
        \bottomrule
    \end{tabular}
    }
\end{table}

\begin{table}[ht]
\centering
\setlength{\tabcolsep}{0.5em}
\caption{\textbf{k-NN Results.} Again, there is no obvious frontrunner.}
\label{tab:knn-results}
\vspace{-0.1in}
\begin{tabular}{@{}l c c c c@{}}
\toprule
\multirow{2}{*}{Method} & \multicolumn{4}{c@{}}{Dataset} \\ \cmidrule{2-5}
{} & \multicolumn{1}{c@{}}{ImageNet} & \multicolumn{1}{c@{}}{Aircraft} & \multicolumn{1}{c@{}}{Flowers} & \multicolumn{1}{c@{}}{NABirds} \\
\midrule
	Supervised & \textbf{73.41} & 31.59 & 77.96 & \textbf{43.25} \\
	BTwins & 62.90 & 31.83 & \textbf{86.18} & 22.29 \\
	DCv2 & 63.70 & \textbf{32.70} & 84.76 & 21.05 \\
	MoCo & 58.59 & 21.39 & 74.53 & 15.40 \\
	SimCLR & 54.57 & 21.21 & 74.78 & 14.03 \\
	SimSiam & 53.66 & 27.39 & 80.01 & 15.18 \\
	SwAV & 61.14 & 28.77 & 82.24 & 15.72 \\
\bottomrule
\end{tabular}

\end{table}

Figure~\ref{fig:all_frozen_results} shows that contrary to claims in some prior work, unsupervised methods do not necessarily struggle on FGVC datasets~\cite{DBLP:journals/corr/abs-2103-16483}.
This is possibly because that work used linear SVMs that were perhaps less accommodating to the ways the unsupervised embeddings tend to be distributed; since our linear evaluation protocol uses batch norm, it is able to better account for this.
Nevertheless, we demonstrate while performance is worse for the two birds datasets and the dogs dataset, Barlow Twins, DeepCluster, and SwAV all outperform supervised pre-training for aircraft, cars, and flowers.
We suggest that significant overlap with ImageNet contributes to part of the gap for performance on CUB and Dogs, which likely confounds those results; nevertheless, NABirds results confirm that unsupervised methods have substantial struggles on that dataset.

\begin{table}[t]
\centering
\setlength{\tabcolsep}{0.5em}
\caption{\textbf{K-Means Results.} Supervised is best for most datasets.}
\label{tab:k-means-results}
\vspace{-0.1in}
\begin{tabular}{@{}l c c c c@{}}
\toprule
\multirow{2}{*}{Method} & \multicolumn{4}{c@{}}{Dataset} \\ \cmidrule{2-5}
{} & \multicolumn{1}{c@{}}{ImageNet} & \multicolumn{1}{c@{}}{Aircraft} & \multicolumn{1}{c@{}}{Flowers} & \multicolumn{1}{c@{}}{NABirds} \\
\midrule
	Supervised & \textbf{58.92} & \textbf{15.69} & 54.97 & \textbf{25.95} \\
	BTwins & 34.88 & 13.20 & \textbf{63.70} & 11.87 \\
	DCv2 & 31.79 & 13.92 & 60.20 & 10.86 \\
	MoCo & 38.30 & 9.84 & 43.34 & 10.75 \\
	SimCLR & 29.78 & 11.16 & 43.99 & 9.08 \\
	SimSiam & 26.20 & 12.66 & 54.51 & 9.53 \\
	SwAV & 28.69 & 12.60 & 56.04 & 9.26 \\
\bottomrule
\end{tabular}

\end{table}

Table~\ref{tab:knn-results} echoes the results in Figure~\ref{fig:all_frozen_results}.
Table~\ref{tab:k-means-results} and Table~\ref{tab:k-means-overcluster-results}, however, show that supervised pre-training dominates the k-means metric, except on Flowers.
It seems clear that pre-training with labels gives supervised learning a strong advantage for k-NN classification and k-means clustering on ImageNet, to the extent that supervised representations even outperform the clustering heads of DeepCluster and SwAV in the overclustering regime on ImageNet.

\begin{table}[t]
\centering
\setlength{\tabcolsep}{0.5em}
\caption{\textbf{K-Means Overclustering Results.} The clustering heads of DeepClusterv2 and SwAV outperform k-means on the learned embeddings of DeepClusterv2 and SwAV.}
\label{tab:k-means-overcluster-results}
\vspace{-0.1in}
\begin{tabular}{@{}l c c c@{}}
\toprule
\multirow{2}{*}{Method} & \multicolumn{3}{c@{}}{ImageNet} \\ \cmidrule{2-4}
{} & \multicolumn{1}{c}{$k=$1000} & \multicolumn{1}{c}{$k=$3000} & \multicolumn{1}{c}{$\Delta$} \\
\midrule
	Supervised K-Means & 58.92 & 65.66 & +6.74 \\
	DCV2 K-Means & 31.79 & 43.02 & +11.23 \\
	DCV2 Clustering Head & n/a & 54.35 & n/a  \\
	SwAV K-Means  & 28.69 & 37.94 & +9.25 \\
	SwAV Clustering Head & n/a & 48.9 & n/a \\
\bottomrule
\end{tabular}

\end{table}

We distill 3 key findings from this.
First, from each of our benchmarks, unsupervised methods are comparable with supervised for generating embeddings for FGVC, and methods like Barlow Twins, DeepCluster, and SwAV seem particularly competitive.
Second, setup matters -- architectural decisions such as the design of the classification head can create subtle biases that favor certain methods, such as the SVM analysis from \cite{DBLP:journals/corr/abs-2103-16483} favoring supervised representations.
Finally, breadth is helpful; each of our benchmarking methods relies on some assumptions, and indirectly evaluates the robustness of the learned embeddings.
Taken together, linear evaluation, k-NN classification, and k-means clustering give a more holistic view of how the representations compare.

\subsection{Uniformity-Tolerance Tradeoff}

\begin{figure}
    \centering
    \includegraphics[width=\linewidth]{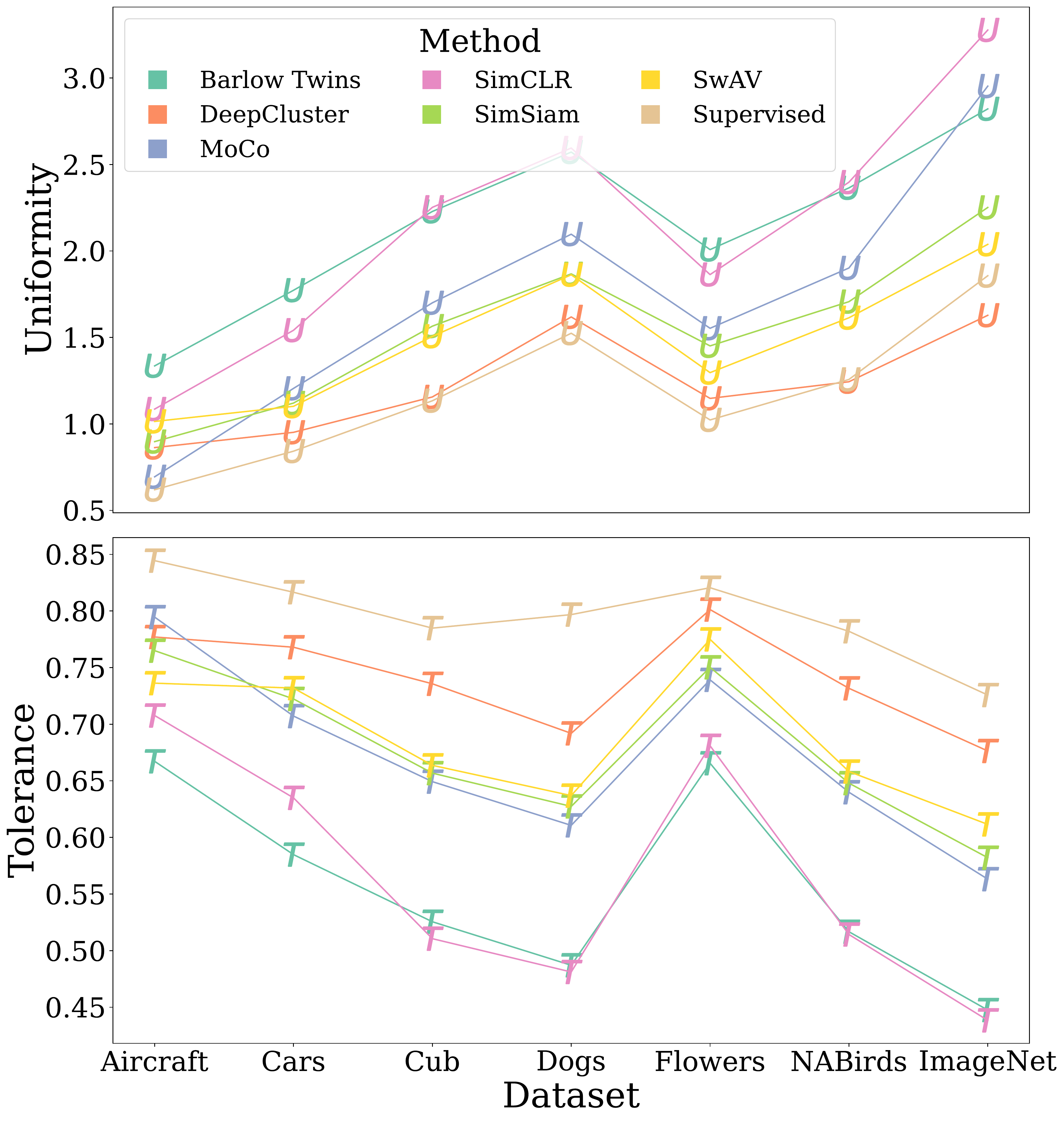}
    \caption{\textbf{Uniformity (U) and Tolerance (T)} on ImageNet and 6 FGVC datasets, with datasets sorted for ascending uniformity. Methods with similar objectives (such as the contrastive methods: Barlow Twins, MoCo, and SimCLR) tend to have similar scores.}
    \label{fig:uniformity_tolerance_curves}
\end{figure}

\begin{figure*}
    \centering
    \includegraphics[width=\linewidth]{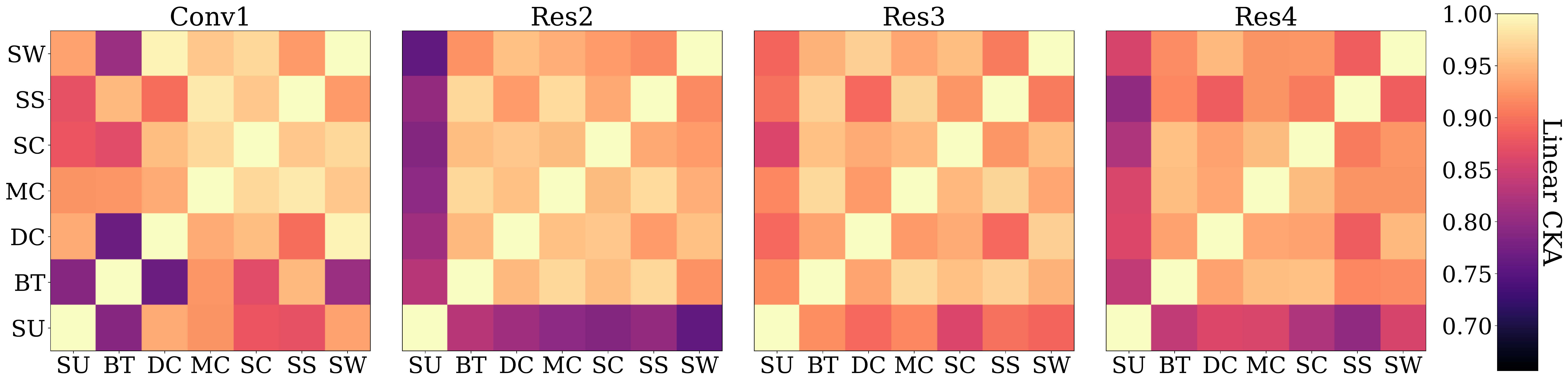}
    \caption{\textbf{Linear CKA for initial convolutional layer and first 3 bottleneck blocks} for validation set of Imagenet. Linear CKA for the last block can be found in Figure~\ref{fig:teaser_star}. In contrast to the final layer, representations are fairly similar in the initial and intermediate blocks.}
    \label{fig:cka_except_last_layer}
    \vspace{-0.1in}
\end{figure*}

High values for uniformity and tolerance are simultaneously desirable, as they indicate favorable distribution of the embeddings on the hypersphere.
Nevertheless, our results in Figure~\ref{fig:uniformity_tolerance_curves} reinforce that in practice, these values have an inverse correlation.
This is because as embeddings are more spread out on the hypersphere in general, they tend to also be more spread out with respect to each ground truth class.
It is perhaps for this reason that supervised pre-training has, in general, the most tolerant and least uniform embeddings.

DeepCluster is similar to supervised for both these metrics.
This is unsurprising when considering the DeepCluster objective: cross-entropy loss for pseudolabels. 
SwAV, with its own clustering objective, exhibits some of these same tendencies, whereas the contrastive methods are the opposite.
This strengthens support for our hypothesis -- unsupervised methods are quite diverse in terms of the distribution of learned embeddings on the hypersphere.

\subsection{Measuring Similarity of Representations}

We consider three main metrics for measuring the similarity between representations: linear CKA, which compares embeddings for any given pair of models, nearest neighbor graph similarity, which compares nearest neighbors for a set of models, and overlap in linear predictions, which compares the predictions of linear classifiers trained on a set of models.
See Figure~\ref{fig:teaser_star} for linear CKA, neighbor similarity, and overlap in linear predictions for the outputs of each ResNet50's final block for ImageNet.
For linear CKA for the other layers we measure, see Figure~\ref{fig:cka_except_last_layer}.
In Table~\ref{tab:linear-overlap-results}, we perform an analysis of linear overlap for groups of models, and leverage ground truth labels to evaluate the uniqueness of the different linear classifiers. 

Results in Figure~\ref{fig:teaser_star} indicate that similar representations tend to have similar neighbors, and classifiers trained on more similar representations tend to make more similar predictions.
The metrics reveal similarities between algorithms with related objectives, such as DeepCluster and SwAV.
More surprisingly, we find more similarity between supervised and unsupervised representations such as DeepCluster than we do between some unsupervised representations, such as MoCo and Barlow Twins. 
Figure~\ref{fig:cka_except_last_layer} simultaneously confirms these findings and those of \cite{grigg2021selfsupervised}, who found that supervised and unsupervised representations diverged the most in the final layer.
We thus extend their hypothesis from SimCLR to additional unsupervised methods, and provide evidence that unsupervised algorithms differ enough from each other to have very different final representations.
This highlights the idea that it is unreasonable to make the ``supervised'' vs. ``unsupervised'' comparisons that are so common in the literature where the ``unsupervised'' is represented by only a couple of algorithms.

\begin{table}[ht]
\centering
\setlength{\tabcolsep}{0.5em}
\caption{\textbf{Results for linear overlap.} 
We examine overlap in predictions for linear classifiers trained on frozen features. 
On the top, we report how many images were predicted correctly by classifiers for both supervised and any unsupervised method, by either, and by neither. 
On the bottom, we compare within unsupervised, considering how many images were uniquely classified correctly by a single linear classifier. 
We find substantial uniqueness for each algorithm, which attests to their complementary nature.
}
\label{tab:linear-overlap-results}
\vspace{-0.1in}
\begin{tabular}{@{}l c c c@{}}
\toprule
\multirow{2}{*}{Method} & \multicolumn{3}{c@{}}{Dataset} \\ \cmidrule{2-4}
{} & \multicolumn{1}{c@{}}{ImageNet} & \multicolumn{1}{c@{}}{Aircraft} & \multicolumn{1}{c@{}}{NABirds} \\
\midrule
	 Sup. and Unsup. & 73.64 & 81.40 & 55.05 \\ 
	 Sup. Only 	 & 2.40 & 0.03 & 6.05 \\ 
	 Unsup. Only 	 & 10.34 & 18.45 & 17.78 \\ 
	 Neither 	 & 13.62 & 0.12 & 21.12 \\ 
\midrule
	 All Unsup. 	 & 58.19 & 80.08 & 30.49 \\ 
	 BTwins Only 	 & 0.97 & 0.24 & 2.46 \\ 
	 DCv2 Only 	 & 1.74 & 0.18 & 4.16 \\ 
	 MoCo Only 	 & 0.69 & 0.09 & 0.87 \\ 
	 SimSiam Only 	 & 0.64 & 0.00 & 0.86 \\ 
	 SwAV Only 	 & 1.74 & 0.21 & 2.80 \\ 
	 No Unsup. 	 & 16.02 & 0.15 & 27.17 \\ 
\bottomrule
\end{tabular}
\vspace{-0.1in}
\end{table}

\begin{figure}
    \centering
    \includegraphics[width=0.9\linewidth]{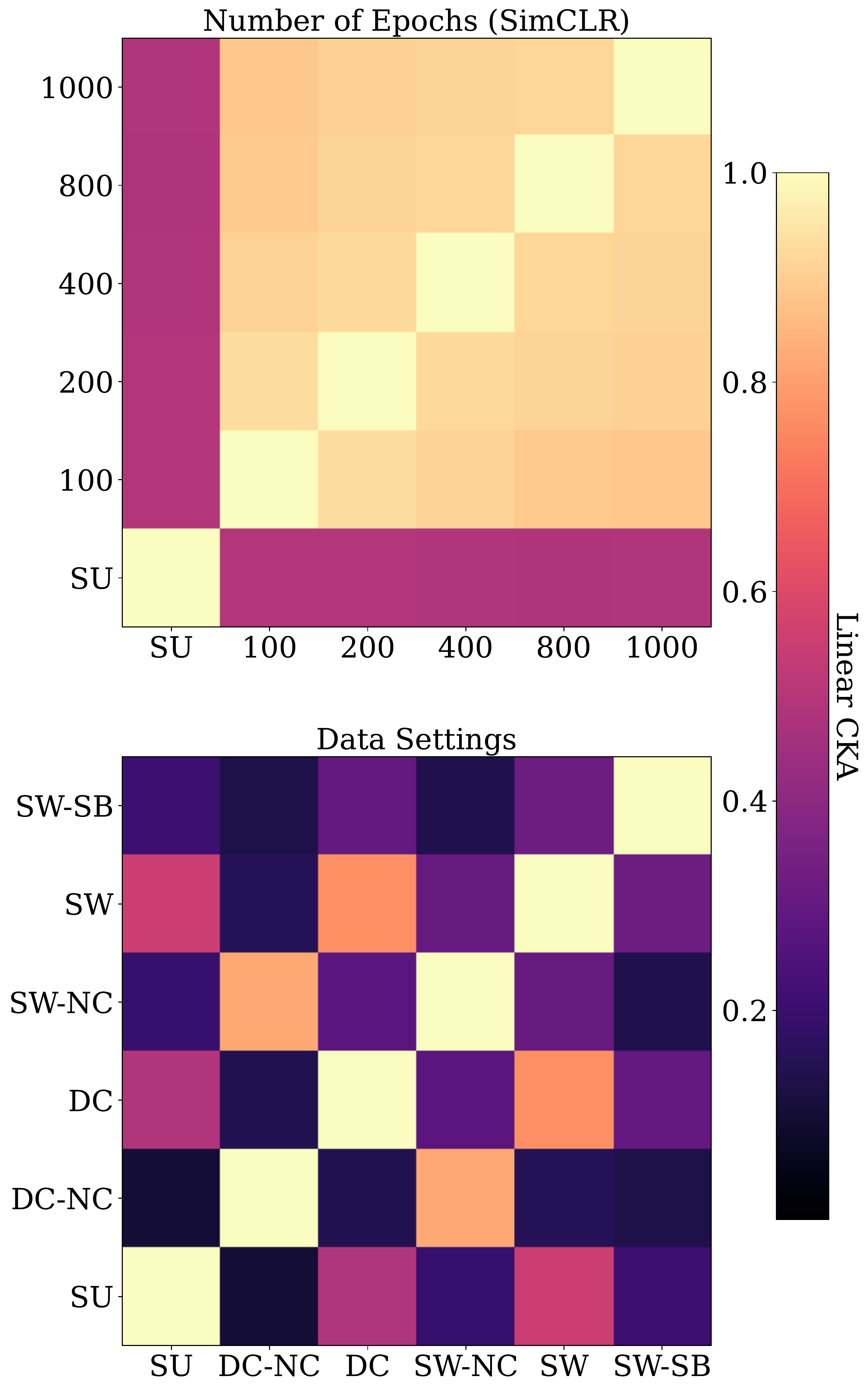}
    \caption{\textbf{Linear CKA for ablation settings} for final residual outputs on the validation set of ImageNet. On the top, we compare supervised (SU) with SimCLR trained for 100, 200, 400, 800, and 1000 epochs. On the bottom, we compare SU with DeepCluster (DC) and SwAV (SW) trained for 400 epochs, as well as with SwAV and DeepCluster trained for 400 epochs with no crops (NC) and SwAV for 400 epochs with a smaller batch size (256).}
    \label{fig:cka_ablation}
\end{figure}

We also compute CKA for unsupervised models under non-default settings, to validate our other findings.
Figure~\ref{fig:cka_ablation}, when compared to Figure~\ref{fig:teaser_star}, shows that training time, at least in the case of SimCLR, induces a comparatively small difference in learned representations, despite the large gaps in performance on benchmarks such as linear evaluation.
While linear evaluation accuracies are different by several percentage points, SimCLR models trained 800 epochs are more similar to the other SimCLR checkpoints than to any other unsupervised algorithm.
Our finding on training time stands in contrast to the results from other settings.
We find details such as cropping strategy and batch size can have a massive impact on the similarity of neural representations, to the extent that SwAV with its full batch size and cropping strategy is more related to DeepCluster with the same settings than to SwAV methods that utilize small batch sizes.
We suggest that future work should probe these effects further, and examine the extent to which the number of crops, batch size, and data augmentations can affect learned representations and downstream applications for various unsupervised algorithms.

\subsection{Augmentation Invariance}
\label{subsec:augmentation-invariance-results}

\begin{figure}
    \centering
    \includegraphics[width=\linewidth]{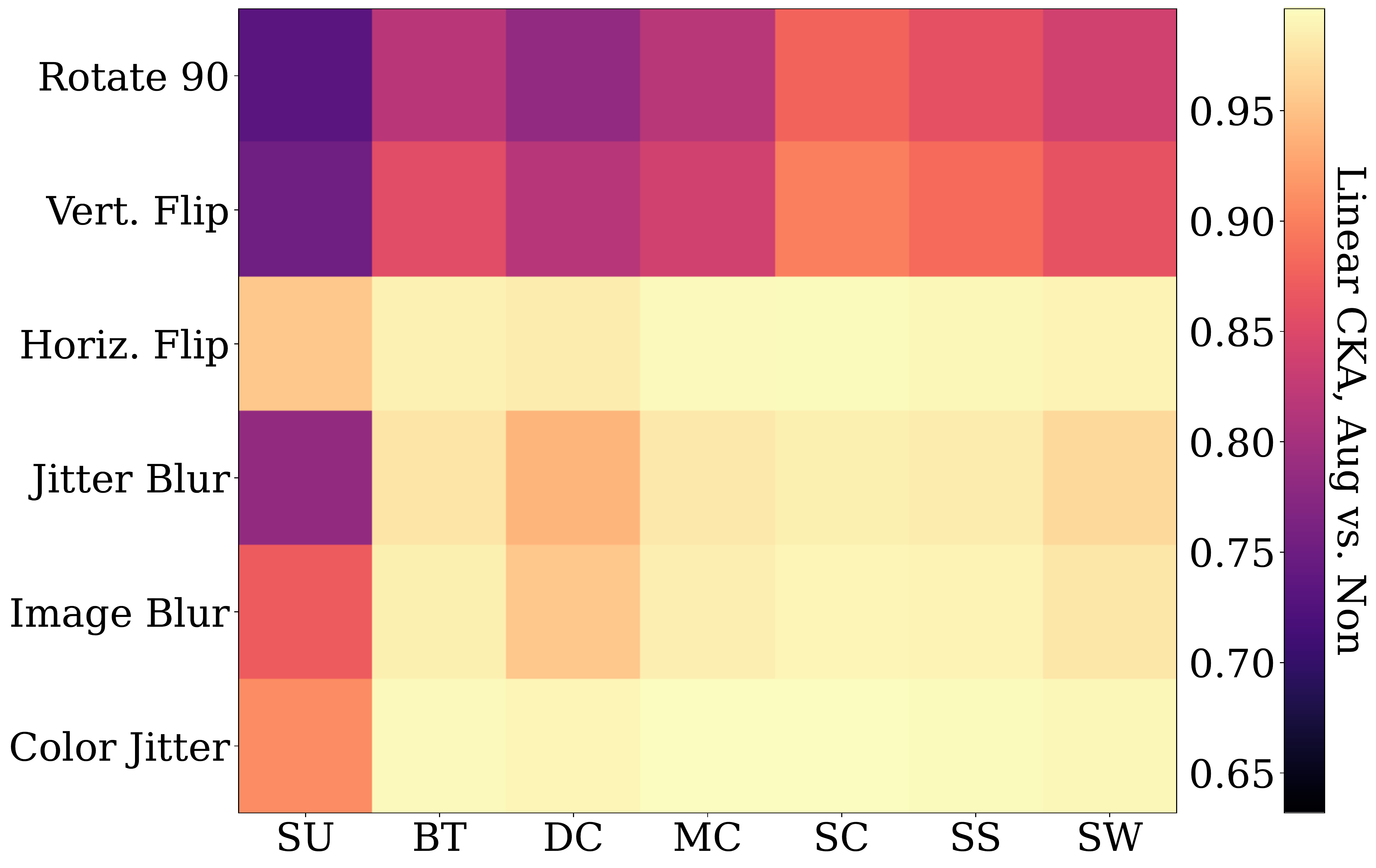}
    \caption{\textbf{Augmentation Invariance}, using Linear CKA for 7 algorithms on ImageNet, for 1 augmentation used in all pretrainings (horizontal flip), 3 used in unsupervised pretraining (color jitter, blurring, and both simultaneously), as well as 2 not used. Unsupervised models exhibit invariance to their training augmentations.}
    \label{fig:augmentation_invariance}
\end{figure}

We use linear CKA to test for augmentation invariance with respect to color jitter, blurring, jitter with blurring, horizontal and vertical flipping, and rotation.
Figure~\ref{fig:augmentation_invariance} provides evidence that, contrary to the conclusions of \cite{DBLP:journals/corr/abs-2007-13916}, and confirming most other prior work, unsupervised algorithms learn representations that are invariant to their training augmentations. 
We note that the invariance is at least somewhat weaker for the clustering algorithms, SwAV and DeepCluster.
Also, the unsupervised methods tend to be somewhat more invariant to augmentations not used at training time, rotation, and vertical flips.
Nevertheless, these experiments suggest that for applications where color is critical, such as bird classification, methods that rely on learned color invariance are destined to underperform.
Future works which seek to mitigate color invariance can leverage our framework as a way to evaluate success.

%% file: sections/5-conclusion.tex
\section{Key Takeaways and Conclusion}
\label{sec:conclusion}



We highlight the following as key takeaways from our findings. 
First, there is no clear ``best'' method. Therefore, it is essential to avoid over-indexing on a given metric, and in the context of applications, representations should be selected to optimize for both downstream data and task.
Second, unsupervised methods share properties that are situationally undesirable, such as robust color invariance. Thus, it is important for future work to develop methods that mitigate certain invariances when necessary, and our CKA-based framework can be utilized to validate their success.
Finally, representations for all algorithms we considered are fairly similar until the last layer, where specialized loss functions or even training settings such as augmentation strategy and batch size can induce learning of substantially different representations. Therefore, it is critical to not assume that one method, such as MoCo or SimCLR, can act as a representative of the field. Additionally, taken in the context of our other findings, we suggest researchers continue to pursue meta-learning, distillation, ensembles, and other approaches that effectively combine different unsupervised algorithms to leverage their complementary nature.

\medskip
\noindent\textbf{Acknowledgements.} This project was partially funded by the DARPA SAIL-ON (W911NF2020009) program, an independent grant from Facebook AI, and Amazon Research Award to AS.

%% file: sections/appendix.tex
\newpage
\appendix

\setcounter{page}{1}

\twocolumn[
\centering
\Large
\textbf{Beyond Supervised vs. Unsupervised: Representative Benchmarking and \\
Analysis of Image Representation Learning} \\ 
\vspace{0.5em}Supplementary Material \\
\vspace{1.0em}
] 
\appendix

\begin{figure*}
    \centering
    \includegraphics[width=\linewidth]{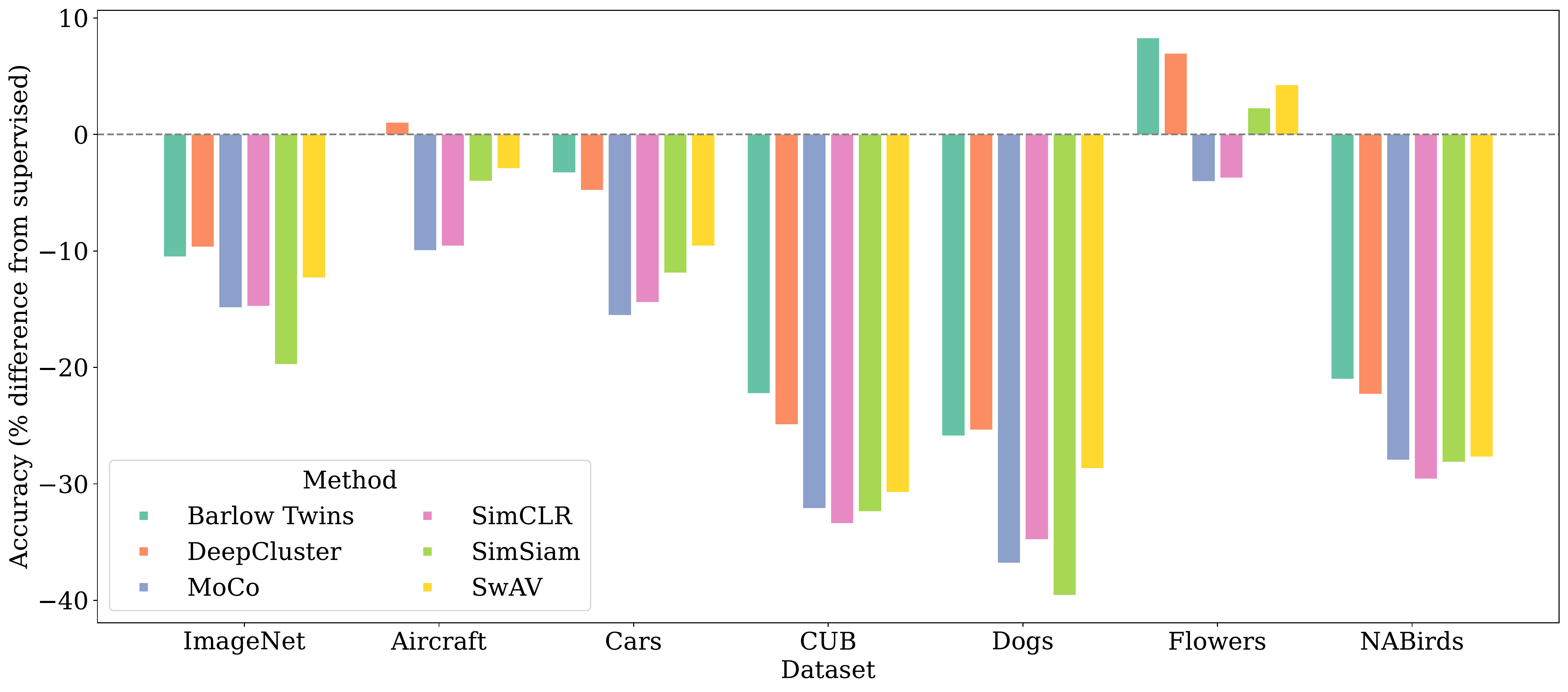}
    \caption{\textbf{k-NN results.}}
    \label{fig:knn_results}
\end{figure*}

\begin{figure*}
    \centering
    \includegraphics[width=\linewidth]{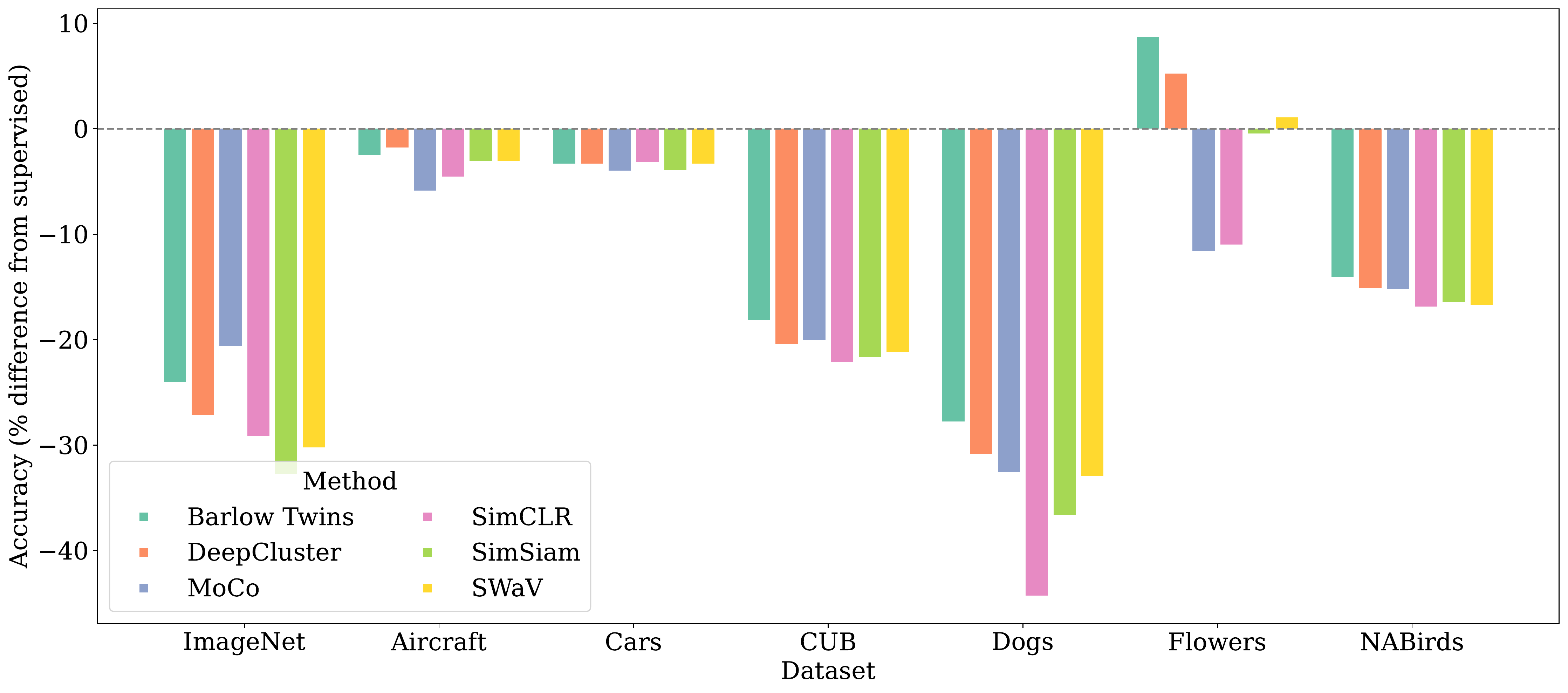}
    \caption{\textbf{K-Means results.}}
    \label{fig:kmeans_results}
\end{figure*}

\section{Code and Assets}

To reproduce our results, please visit our repository at https://github.com/mgwillia/unsupervised-analysis.
Where specified (see our repository for details), we use code from VISSL (https://github.com/facebookresearch/vissl/) and SCAN (https://github.com/wvangansbeke/Unsupervised-Classification). 
These repositories have an MIT License and Creative Commons License, respectively.

\section{k-NN Details}

For k-NN classification, we use the VISSL defaults for ImageNet: 200 neighbors.
For the FGVC datasets, there are too few images per class to use this approach.
Instead, we try values in the set $\{0, 5, 10, 15, 20, 25, 30, 35, 40, 45, 50\}$ and choose whichever value maximizes accuracy.

\section{More Benchmark Results}

Here, we give an expanded look at our benchmarks.
%
%
Table~\ref{tab:linear-eval-full-results} complements Figure 3 by providing the same data, in tabular form.
Figures \ref{fig:knn_results} and \ref{fig:kmeans_results} along with Tables \ref{tab:knn-full-results} and \ref{tab:kmeans-full-results} do the same for k-NN and k-means, offering an expansion of the results shown in Tables 2 and 3.

\begin{table*}[ht]
\centering
\setlength{\tabcolsep}{0.5em}
\caption{\textbf{Linear evaluation results.}}
\begin{tabular}{@{}l c c c c c c c@{}}
\toprule
\multirow{2}{*}{Method} & \multicolumn{7}{c}{Dataset} \\ \cmidrule{2-8}
{} & \multicolumn{1}{c@{}}{ImageNet} & \multicolumn{1}{c@{}}{Aircraft} & \multicolumn{1}{c@{}}{Cars} & \multicolumn{1}{c@{}}{CUB} & \multicolumn{1}{c@{}}{Dogs} & \multicolumn{1}{c@{}}{Flowers} & \multicolumn{1}{c@{}}{NABirds} \\
\midrule
	Supervised & \textbf{76.04} & 48.05 & 57.72 & \textbf{70.57} & \textbf{88.92} & 91.30 & \textbf{61.10} \\
	Barlow Twins & 71.78 & \textbf{58.51} & 65.30 & 63.98 & 74.35 & 94.21 & 54.59 \\
	DeepCluster & 75.19 & 58.36 & \textbf{67.76} & 69.82 & 77.80 & \textbf{94.46} & 59.89 \\
	MoCo & 69.95 & 49.88 & 49.68 & 51.95 & 67.93 & 91.14 & 43.51 \\
	SimCLR & 68.95 & 44.39 & 45.00 & 47.36 & 65.84 & 88.90 & 38.43 \\
	SimSiam & 67.89 & 52.22 & 53.34 & 52.55 & 66.50 & 91.86 & 44.18 \\
	SwAV & 74.87 & 55.73 & 61.95 & 65.10 & 75.99 & 93.97 & 56.52 \\
\bottomrule
\end{tabular}
\label{tab:linear-eval-full-results}
\end{table*}

\begin{table*}[ht]
\centering
\setlength{\tabcolsep}{0.5em}
\caption{\textbf{k-NN results.}}
\begin{tabular}{@{}l c c c c c c c@{}}
\toprule
\multirow{2}{*}{Method} & \multicolumn{7}{c@{}}{Dataset} \\ \cmidrule{2-8}
{} & \multicolumn{1}{c@{}}{ImageNet} & \multicolumn{1}{c@{}}{Aircraft} & \multicolumn{1}{c@{}}{Cars} & \multicolumn{1}{c@{}}{CUB} & \multicolumn{1}{c@{}}{Dogs} & \multicolumn{1}{c@{}}{Flowers} & \multicolumn{1}{c@{}}{NABirds} \\
\midrule
	Supervised & 73.41 & 31.59 & 30.16 & 56.63 & 88.38 & 77.96 & 43.25 \\
	Barlow Twins & 62.90 & 31.83 & 26.94 & 34.41 & 62.53 & 86.18 & 22.29 \\
	DeepCluster & 63.70 & 32.70 & 25.48 & 31.74 & 62.97 & 84.76 & 21.05 \\
	MoCo & 58.59 & 21.39 & 14.64 & 24.35 & 51.60 & 74.53 & 15.40 \\
	SimCLR & 54.57 & 21.21 & 14.74 & 23.21 & 49.63 & 74.78 & 14.03 \\
	SimSiam & 53.66 & 27.39 & 18.41 & 24.20 & 48.97 & 80.01 & 15.18 \\
	SwAV & 61.14 & 28.77 & 20.84 & 25.75 & 59.87 & 82.24 & 15.72 \\
\bottomrule
\end{tabular}
\label{tab:knn-full-results}
\end{table*}

\begin{table*}[ht]
\centering
\setlength{\tabcolsep}{0.5em}
\caption{\textbf{K-Means results.}}
\begin{tabular}{@{}l c c c c c c c@{}}
\toprule
\multirow{2}{*}{Method} & \multicolumn{7}{c@{}}{Dataset} \\ \cmidrule{2-8}
{} & \multicolumn{1}{c@{}}{ImageNet} & \multicolumn{1}{c@{}}{Aircraft} & \multicolumn{1}{c@{}}{Cars} & \multicolumn{1}{c@{}}{CUB} & \multicolumn{1}{c@{}}{Dogs} & \multicolumn{1}{c@{}}{Flowers} & \multicolumn{1}{c@{}}{NABirds} \\
\midrule
	Supervised & 58.92 & 15.69 & 11.95 & 35.23 & 53.69 & 54.97 & 25.95 \\
	DeepCluster & 31.79 & 13.92 & 8.66 & 14.81 & 22.84 & 60.20 & 10.86 \\
	MoCo & 38.30 & 9.84 & 7.98 & 15.21 & 21.10 & 43.34 & 10.75 \\
	Barlow Twins & 34.88 & 13.20 & 8.63 & 17.07 & 25.94 & 63.70 & 11.87 \\
	SimCLR & 29.78 & 11.16 & 8.80 & 13.07 & 9.41 & 43.99 & 9.08 \\
	SimSiam & 26.20 & 12.66 & 8.03 & 13.57 & 17.07 & 54.51 & 9.53 \\
	SwAV & 28.69 & 12.60 & 8.66 & 14.05 & 20.79 & 56.04 & 9.26 \\
\bottomrule
\end{tabular}
\label{tab:kmeans-full-results}
\end{table*}

We verify claims we make about the SimCLR models from the main paper.
Specifically, we say that training time has a significant impact on results, while not changing the representations substantially (see Figure 6).
Table~\ref{tab:sclr-epoch-lin-eval} offers evidence supporting our claim.

\begin{table*}[ht]
\centering
\setlength{\tabcolsep}{0.5em}
\caption{\textbf{Linear Evaluation for SimCLR} with varying training time.}
\begin{tabular}{@{}l c c c c c c c@{}}
\toprule
\multirow{2}{*}{Method} & \multicolumn{7}{c@{}}{Dataset} \\ \cmidrule{2-8}
{} & \multicolumn{1}{c@{}}{ImageNet} & \multicolumn{1}{c@{}}{Aircraft} & \multicolumn{1}{c@{}}{Cars} & \multicolumn{1}{c@{}}{CUB} & \multicolumn{1}{c@{}}{Dogs} & \multicolumn{1}{c@{}}{Flowers} & \multicolumn{1}{c@{}}{NABirds} \\
\midrule
	100 Epochs & 64.76 & 44.81 & 44.67 & 43.17 & 60.44 & 88.72 & 34.34 \\
	200 Epochs & 66.92 & \textbf{45.56} & 46.31 & 46.05 & 62.48 & \textbf{89.39} & 36.90 \\
	400 Epochs & 67.93 & 44.84 & \textbf{46.36} & 46.00 & 64.35 & 89.08 & 37.23 \\
	800 Epochs & \textbf{68.95} & 44.39 & 45.00 & \textbf{47.36} & 65.84 & 88.90 & \textbf{38.43} \\
	1000 Epochs & 64.57 & 45.26 & 44.55 & 46.93 & \textbf{66.25} & 88.57 & 37.93 \\
\bottomrule
\end{tabular}
\label{tab:sclr-epoch-lin-eval}
\end{table*}

%% file: main.bbl
\begin{thebibliography}{10}\itemsep=-1pt

\bibitem{DBLP:journals/corr/abs-2012-00868}
Srikar Appalaraju, Yi Zhu, Yusheng Xie, and Istv{\'{a}}n Feh{\'{e}}rv{\'{a}}ri.
\newblock Towards good practices in self-supervised representation learning.
\newblock {\em CoRR}, abs/2012.00868, 2020.

\bibitem{arthur2006k}
David Arthur and Sergei Vassilvitskii.
\newblock k-means++: The advantages of careful seeding.
\newblock Technical report, Stanford, 2006.

\bibitem{asano2019self}
Yuki~Markus Asano, Christian Rupprecht, and Andrea Vedaldi.
\newblock Self-labelling via simultaneous clustering and representation
  learning.
\newblock {\em arXiv preprint arXiv:1911.05371}, 2019.

\bibitem{bachman2019learning}
Philip Bachman, R~Devon Hjelm, and William Buchwalter.
\newblock Learning representations by maximizing mutual information across
  views.
\newblock {\em arXiv preprint arXiv:1906.00910}, 2019.

\bibitem{brock2018large}
Andrew Brock, Jeff Donahue, and Karen Simonyan.
\newblock Large scale gan training for high fidelity natural image synthesis.
\newblock {\em arXiv preprint arXiv:1809.11096}, 2018.

\bibitem{caron2018deep}
Mathilde Caron, Piotr Bojanowski, Armand Joulin, and Matthijs Douze.
\newblock Deep clustering for unsupervised learning of visual features.
\newblock In {\em Proceedings of the European Conference on Computer Vision
  (ECCV)}, pages 132--149, 2018.

\bibitem{caron2019unsupervised}
Mathilde Caron, Piotr Bojanowski, Julien Mairal, and Armand Joulin.
\newblock Unsupervised pre-training of image features on non-curated data.
\newblock In {\em Proceedings of the IEEE/CVF International Conference on
  Computer Vision}, pages 2959--2968, 2019.

\bibitem{caron2020unsupervised}
Mathilde Caron, Ishan Misra, Julien Mairal, Priya Goyal, Piotr Bojanowski, and
  Armand Joulin.
\newblock Unsupervised learning of visual features by contrasting cluster
  assignments.
\newblock {\em arXiv preprint arXiv:2006.09882}, 2020.

\bibitem{caron2021emerging}
Mathilde Caron, Hugo Touvron, Ishan Misra, Herv{\'e} J{\'e}gou, Julien Mairal,
  Piotr Bojanowski, and Armand Joulin.
\newblock Emerging properties in self-supervised vision transformers.
\newblock {\em arXiv preprint arXiv:2104.14294}, 2021.

\bibitem{pmlr-v119-chen20j}
Ting Chen, Simon Kornblith, Mohammad Norouzi, and Geoffrey Hinton.
\newblock A simple framework for contrastive learning of visual
  representations.
\newblock In Hal~Daumé III and Aarti Singh, editors, {\em Proceedings of the
  37th International Conference on Machine Learning}, volume 119 of {\em
  Proceedings of Machine Learning Research}, pages 1597--1607. PMLR, 13--18 Jul
  2020.

\bibitem{chen2020simple}
Ting Chen, Simon Kornblith, Mohammad Norouzi, and Geoffrey Hinton.
\newblock A simple framework for contrastive learning of visual
  representations.
\newblock In {\em International conference on machine learning}, pages
  1597--1607. PMLR, 2020.

\bibitem{chen2020improved}
Xinlei Chen, Haoqi Fan, Ross Girshick, and Kaiming He.
\newblock Improved baselines with momentum contrastive learning.
\newblock {\em arXiv preprint arXiv:2003.04297}, 2020.

\bibitem{chen2021exploring}
Xinlei Chen and Kaiming He.
\newblock Exploring simple siamese representation learning.
\newblock In {\em Proceedings of the IEEE/CVF Conference on Computer Vision and
  Pattern Recognition}, pages 15750--15758, 2021.

\bibitem{DBLP:journals/corr/abs-2105-05837}
Elijah Cole, Xuan Yang, Kimberly Wilber, Oisin~Mac Aodha, and Serge~J.
  Belongie.
\newblock When does contrastive visual representation learning work?
\newblock {\em CoRR}, abs/2105.05837, 2021.

\bibitem{JMLR:v13:cortes12a}
Corinna Cortes, Mehryar Mohri, and Afshin Rostamizadeh.
\newblock Algorithms for learning kernels based on centered alignment.
\newblock {\em Journal of Machine Learning Research}, 13(28):795--828, 2012.

\bibitem{5206848}
Jia Deng, Wei Dong, Richard Socher, Li-Jia Li, Kai Li, and Li Fei-Fei.
\newblock Imagenet: A large-scale hierarchical image database.
\newblock In {\em 2009 IEEE Conference on Computer Vision and Pattern
  Recognition}, pages 248--255, 2009.

\bibitem{doersch2015unsupervised}
Carl Doersch, Abhinav Gupta, and Alexei~A Efros.
\newblock Unsupervised visual representation learning by context prediction.
\newblock In {\em Proceedings of the IEEE international conference on computer
  vision}, pages 1422--1430, 2015.

\bibitem{doersch2017multi}
Carl Doersch and Andrew Zisserman.
\newblock Multi-task self-supervised visual learning.
\newblock In {\em Proceedings of the IEEE International Conference on Computer
  Vision}, pages 2051--2060, 2017.

\bibitem{donahue2016adversarial}
Jeff Donahue, Philipp Kr{\"a}henb{\"u}hl, and Trevor Darrell.
\newblock Adversarial feature learning.
\newblock {\em arXiv preprint arXiv:1605.09782}, 2016.

\bibitem{donahue2019large}
Jeff Donahue and Karen Simonyan.
\newblock Large scale adversarial representation learning.
\newblock {\em arXiv preprint arXiv:1907.02544}, 2019.

\bibitem{DBLP:journals/corr/abs-2011-13377}
Linus Ericsson, Henry Gouk, and Timothy~M. Hospedales.
\newblock How well do self-supervised models transfer?
\newblock {\em CoRR}, abs/2011.13377, 2020.

\bibitem{DBLP:journals/corr/abs-2106-05967}
Wouter~Van Gansbeke, Simon Vandenhende, Stamatios Georgoulis, and Luc~Van Gool.
\newblock Revisiting contrastive methods for unsupervised learning of visual
  representations.
\newblock {\em CoRR}, abs/2106.05967, 2021.

\bibitem{gidaris2018unsupervised}
Spyros Gidaris, Praveer Singh, and Nikos Komodakis.
\newblock Unsupervised representation learning by predicting image rotations.
\newblock {\em arXiv preprint arXiv:1803.07728}, 2018.

\bibitem{goyal2021vissl}
Priya Goyal, Quentin Duval, Jeremy Reizenstein, Matthew Leavitt, Min Xu,
  Benjamin Lefaudeux, Mannat Singh, Vinicius Reis, Mathilde Caron, Piotr
  Bojanowski, Armand Joulin, and Ishan Misra.
\newblock Vissl.
\newblock \url{https://github.com/facebookresearch/vissl}, 2021.

\bibitem{gretton2007kernel}
Arthur Gretton, Kenji Fukumizu, Choon~Hui Teo, Le Song, Bernhard Sch{\"o}lkopf,
  Alexander~J Smola, et~al.
\newblock A kernel statistical test of independence.
\newblock In {\em Nips}, volume~20, pages 585--592. Citeseer, 2007.

\bibitem{grigg2021selfsupervised}
Tom~George Grigg, Dan Busbridge, Jason Ramapuram, and Russ Webb.
\newblock Do self-supervised and supervised methods learn similar visual
  representations?, 2021.

\bibitem{gutmann2010noise}
Michael Gutmann and Aapo Hyv{\"a}rinen.
\newblock Noise-contrastive estimation: A new estimation principle for
  unnormalized statistical models.
\newblock In {\em Proceedings of the thirteenth international conference on
  artificial intelligence and statistics}, pages 297--304. JMLR Workshop and
  Conference Proceedings, 2010.

\bibitem{hadsell2006dimensionality}
Raia Hadsell, Sumit Chopra, and Yann LeCun.
\newblock Dimensionality reduction by learning an invariant mapping.
\newblock In {\em 2006 IEEE Computer Society Conference on Computer Vision and
  Pattern Recognition (CVPR'06)}, volume~2, pages 1735--1742. IEEE, 2006.

\bibitem{DBLP:journals/corr/abs-1911-05722}
Kaiming He, Haoqi Fan, Yuxin Wu, Saining Xie, and Ross~B. Girshick.
\newblock Momentum contrast for unsupervised visual representation learning.
\newblock {\em CoRR}, abs/1911.05722, 2019.

\bibitem{he2015deep}
Kaiming He, Xiangyu Zhang, Shaoqing Ren, and Jian Sun.
\newblock Deep residual learning for image recognition, 2015.

\bibitem{henaff2020data}
Olivier Henaff.
\newblock Data-efficient image recognition with contrastive predictive coding.
\newblock In {\em International Conference on Machine Learning}, pages
  4182--4192. PMLR, 2020.

\bibitem{hjelm2018learning}
R~Devon Hjelm, Alex Fedorov, Samuel Lavoie-Marchildon, Karan Grewal, Phil
  Bachman, Adam Trischler, and Yoshua Bengio.
\newblock Learning deep representations by mutual information estimation and
  maximization.
\newblock {\em arXiv preprint arXiv:1808.06670}, 2018.

\bibitem{DBLP:journals/corr/abs-2103-16483}
Grant~Van Horn, Elijah Cole, Sara Beery, Kimberly Wilber, Serge~J. Belongie,
  and Oisin~Mac Aodha.
\newblock Benchmarking representation learning for natural world image
  collections.
\newblock {\em CoRR}, abs/2103.16483, 2021.

\bibitem{KhoslaYaoJayadevaprakashFeiFei_FGVC2011}
Aditya Khosla, Nityananda Jayadevaprakash, Bangpeng Yao, and Li Fei-Fei.
\newblock Novel dataset for fine-grained image categorization.
\newblock In {\em First Workshop on Fine-Grained Visual Categorization, IEEE
  Conference on Computer Vision and Pattern Recognition}, Colorado Springs, CO,
  June 2011.

\bibitem{kingma2013auto}
Diederik~P Kingma and Max Welling.
\newblock Auto-encoding variational bayes.
\newblock {\em arXiv preprint arXiv:1312.6114}, 2013.

\bibitem{pmlr-v97-kornblith19a}
Simon Kornblith, Mohammad Norouzi, Honglak Lee, and Geoffrey Hinton.
\newblock Similarity of neural network representations revisited.
\newblock In Kamalika Chaudhuri and Ruslan Salakhutdinov, editors, {\em
  Proceedings of the 36th International Conference on Machine Learning},
  volume~97 of {\em Proceedings of Machine Learning Research}, pages
  3519--3529. PMLR, 09--15 Jun 2019.

\bibitem{Kotar_2021_ICCV}
Klemen Kotar, Gabriel Ilharco, Ludwig Schmidt, Kiana Ehsani, and Roozbeh
  Mottaghi.
\newblock Contrasting contrastive self-supervised representation learning
  pipelines.
\newblock In {\em Proceedings of the IEEE/CVF International Conference on
  Computer Vision (ICCV)}, pages 9949--9959, October 2021.

\bibitem{KrauseStarkDengFei-Fei_3DRR2013}
Jonathan Krause, Michael Stark, Jia Deng, and Li Fei-Fei.
\newblock 3d object representations for fine-grained categorization.
\newblock In {\em 4th International IEEE Workshop on 3D Representation and
  Recognition (3dRR-13)}, Sydney, Australia, 2013.

\bibitem{larsson2016learning}
Gustav Larsson, Michael Maire, and Gregory Shakhnarovich.
\newblock Learning representations for automatic colorization.
\newblock In {\em European conference on computer vision}, pages 577--593.
  Springer, 2016.

\bibitem{li2021efficient}
Chunyuan Li, Jianwei Yang, Pengchuan Zhang, Mei Gao, Bin Xiao, Xiyang Dai, Lu
  Yuan, and Jianfeng Gao.
\newblock Efficient self-supervised vision transformers for representation
  learning.
\newblock {\em arXiv preprint arXiv:2106.09785}, 2021.

\bibitem{maji13fine-grained}
S. Maji, J. Kannala, E. Rahtu, M. Blaschko, and A. Vedaldi.
\newblock Fine-grained visual classification of aircraft.
\newblock Technical report, 2013.

\bibitem{misra2020self}
Ishan Misra and Laurens van~der Maaten.
\newblock Self-supervised learning of pretext-invariant representations.
\newblock In {\em Proceedings of the IEEE/CVF Conference on Computer Vision and
  Pattern Recognition}, pages 6707--6717, 2020.

\bibitem{Nilsback08}
Maria-Elena Nilsback and Andrew Zisserman.
\newblock Automated flower classification over a large number of classes.
\newblock In {\em Indian Conference on Computer Vision, Graphics and Image
  Processing}, Dec 2008.

\bibitem{noroozi2016unsupervised}
Mehdi Noroozi and Paolo Favaro.
\newblock Unsupervised learning of visual representations by solving jigsaw
  puzzles.
\newblock In {\em European conference on computer vision}, pages 69--84.
  Springer, 2016.

\bibitem{oord2018representation}
Aaron van~den Oord, Yazhe Li, and Oriol Vinyals.
\newblock Representation learning with contrastive predictive coding.
\newblock {\em arXiv preprint arXiv:1807.03748}, 2018.

\bibitem{pathak2016context}
Deepak Pathak, Philipp Krahenbuhl, Jeff Donahue, Trevor Darrell, and Alexei~A
  Efros.
\newblock Context encoders: Feature learning by inpainting.
\newblock In {\em Proceedings of the IEEE conference on computer vision and
  pattern recognition}, pages 2536--2544, 2016.

\bibitem{DBLP:journals/corr/abs-2007-13916}
Senthil Purushwalkam and Abhinav Gupta.
\newblock Demystifying contrastive self-supervised learning: Invariances,
  augmentations and dataset biases.
\newblock {\em CoRR}, abs/2007.13916, 2020.

\bibitem{tian2020contrastive}
Yonglong Tian, Dilip Krishnan, and Phillip Isola.
\newblock Contrastive multiview coding.
\newblock In {\em Computer Vision--ECCV 2020: 16th European Conference,
  Glasgow, UK, August 23--28, 2020, Proceedings, Part XI 16}, pages 776--794.
  Springer, 2020.

\bibitem{van2020scan}
Wouter Van~Gansbeke, Simon Vandenhende, Stamatios Georgoulis, Marc Proesmans,
  and Luc Van~Gool.
\newblock Scan: Learning to classify images without labels.
\newblock In {\em European Conference on Computer Vision}, pages 268--285.
  Springer, 2020.

\bibitem{7298658}
Grant Van~Horn, Steve Branson, Ryan Farrell, Scott Haber, Jessie Barry, Panos
  Ipeirotis, Pietro Perona, and Serge Belongie.
\newblock Building a bird recognition app and large scale dataset with citizen
  scientists: The fine print in fine-grained dataset collection.
\newblock In {\em 2015 IEEE Conference on Computer Vision and Pattern
  Recognition (CVPR)}, pages 595--604, 2015.

\bibitem{vincent2008extracting}
Pascal Vincent, Hugo Larochelle, Yoshua Bengio, and Pierre-Antoine Manzagol.
\newblock Extracting and composing robust features with denoising autoencoders.
\newblock In {\em Proceedings of the 25th international conference on Machine
  learning}, pages 1096--1103, 2008.

\bibitem{WahCUB_200_2011}
C. Wah, S. Branson, P. Welinder, P. Perona, and S. Belongie.
\newblock {The Caltech-UCSD Birds-200-2011 Dataset}.
\newblock Technical Report CNS-TR-2011-001, California Institute of Technology,
  2011.

\bibitem{Wang_2021_CVPR}
Feng Wang and Huaping Liu.
\newblock Understanding the behaviour of contrastive loss.
\newblock In {\em Proceedings of the IEEE/CVF Conference on Computer Vision and
  Pattern Recognition (CVPR)}, pages 2495--2504, June 2021.

\bibitem{DBLP:journals/corr/abs-2005-10242}
Tongzhou Wang and Phillip Isola.
\newblock Understanding contrastive representation learning through alignment
  and uniformity on the hypersphere.
\newblock {\em CoRR}, abs/2005.10242, 2020.

\bibitem{Wu_2018_CVPR}
Zhirong Wu, Yuanjun Xiong, Stella~X. Yu, and Dahua Lin.
\newblock Unsupervised feature learning via non-parametric instance
  discrimination.
\newblock In {\em Proceedings of the IEEE Conference on Computer Vision and
  Pattern Recognition (CVPR)}, June 2018.

\bibitem{xiao2021what}
Tete Xiao, Xiaolong Wang, Alexei~A Efros, and Trevor Darrell.
\newblock What should not be contrastive in contrastive learning.
\newblock In {\em International Conference on Learning Representations}, 2021.

\bibitem{xie2021self}
Zhenda Xie, Yutong Lin, Zhuliang Yao, Zheng Zhang, Qi Dai, Yue Cao, and Han Hu.
\newblock Self-supervised learning with swin transformers.
\newblock {\em arXiv preprint arXiv:2105.04553}, 2021.

\bibitem{Ye_2019_CVPR}
Mang Ye, Xu Zhang, Pong~C. Yuen, and Shih-Fu Chang.
\newblock Unsupervised embedding learning via invariant and spreading instance
  feature.
\newblock In {\em Proceedings of the IEEE/CVF Conference on Computer Vision and
  Pattern Recognition (CVPR)}, June 2019.

\bibitem{DBLP:journals/corr/abs-2103-03230}
Jure Zbontar, Li Jing, Ishan Misra, Yann LeCun, and St{\'{e}}phane Deny.
\newblock Barlow twins: Self-supervised learning via redundancy reduction.
\newblock {\em CoRR}, abs/2103.03230, 2021.

\bibitem{zhang2016colorful}
Richard Zhang, Phillip Isola, and Alexei~A Efros.
\newblock Colorful image colorization.
\newblock In {\em European conference on computer vision}, pages 649--666.
  Springer, 2016.

\end{thebibliography}
